\pdfoutput=1
\documentclass[11pt]{article}
\usepackage[final]{acl}

\usepackage[T1]{fontenc}
\usepackage[utf8]{inputenc}

\usepackage{times}
\usepackage{latexsym}
\usepackage{microtype}
\usepackage{xcolor}

\usepackage{graphicx}
\graphicspath{{figure/}}
\usepackage{booktabs}
\usepackage{multirow}
\usepackage{makecell}

\usepackage{amsmath}

\usepackage{algorithm}
\usepackage{algpseudocode}

\usepackage{subcaption}

\usepackage[most]{tcolorbox}
\usepackage{listings}

\usepackage{fontawesome}

\definecolor{babyblue}{HTML}{89CFF0}
\newtcbox{\highlight}[1][babyblue]{on line, colback=#1!20, colframe=#1!80!black,
  boxrule=0pt, arc=3pt, boxsep=0pt, left=1pt, right=1pt, top=2pt, bottom=2pt,
  fontupper=\ttfamily}

\title{Visual Self-Fulfilling Alignment: Shaping Safety-Oriented Personas via Threat-Related Images}

\author{
  \textbf{Qishun Yang\textsuperscript{1,2,4,*}} \quad
  \textbf{Shu Yang\textsuperscript{1,2,*}\textsuperscript{†}} \quad
  \textbf{Lijie Hu\textsuperscript{3}} \quad
  \textbf{Di Wang\textsuperscript{1,2}\textsuperscript{†}} \\
  \textsuperscript{1}King Abdullah University of Science and Technology\\
  \textsuperscript{2}Provable Responsible AI and Data Analytics Lab \\
  \textsuperscript{3}Mohamed bin Zayed University of Artificial Intelligence \\
  \textsuperscript{4}China University of Petroleum-Beijing at Karamay \\
}

\begin{document}
\maketitle

\begingroup
\renewcommand{\thefootnote}{\fnsymbol{footnote}}
\footnotetext[1]{Equal contribution and shared co-first authorship.}
\footnotetext[2]{Corresponding author.}
\endgroup
\setcounter{footnote}{0}

\begin{abstract}
Multimodal large language models (MLLMs) face safety misalignment, where visual inputs enable harmful outputs. To address this, existing methods require explicit safety labels or contrastive data; yet, threat-related concepts are concrete and visually depictable, while safety concepts, like helpfulness, are abstract and lack visual referents. Inspired by the Self-Fulfilling mechanism underlying emergent misalignment, we propose Visual Self-Fulfilling Alignment (VSFA). VSFA fine-tunes vision-language models (VLMs) on neutral VQA tasks constructed around threat-related images, without any safety labels. Through repeated exposure to threat-related visual content, models internalize the implicit semantics of vigilance and caution, shaping safety-oriented personas. Experiments across multiple VLMs and safety benchmarks demonstrate that VSFA reduces the attack success rate, improves response quality, and mitigates over-refusal while preserving general capabilities. Our work extends the self-fulfilling mechanism from text to visual modalities, offering a label-free approach to VLMs alignment. Code is available at \url{https://github.com/qazwsx123456123/VSFA}.
\end{abstract}

\section{Introduction}
\label{sec:intro}

\noindent Multimodal large language models (MLLMs) integrate vision and language capabilities, demonstrating strong performance across diverse applications~\cite{li2025towards, jiang2025tc, veagle2024, jiang2025hykge,zhang2026stackplanner}. These models handle tasks ranging from visual question answering (VQA) to complex reasoning with multimodal content~\cite{dynamicllava2024, zhou2025flattery,yang2025d}. Many widely-used MLLMs are built upon large language models (LLMs) that have been aligned with human values through textual training, such as LLaVA~\cite{llava2023} and Qwen2-VL~\cite{wang2024qwen2vl}. However, visual inputs introduce vulnerabilities absent in text-only systems, such as adversarial perturbation attacks, where imperceptible perturbations added to images cause abnormal model behavior~\cite{tang2025safetyreminder}. Furthermore, integrating visual modality creates a modality gap, where images and text are embedded separately in the representation space~\cite{liang2022mindgap}. This separation weakens safety awareness. Harmful images can conceal and intensify dangerous intent within textual queries~\cite{li2024images}, and visual information leakage further circumvents textual safety filters~\cite{vlsbench2024}. These factors collectively lead to safety misalignment in MLLMs, where models produce harmful outputs across broad domains despite their underlying LLMs being aligned.

\begin{figure*}[t]
    \centering
    \includegraphics[width=0.88\textwidth]{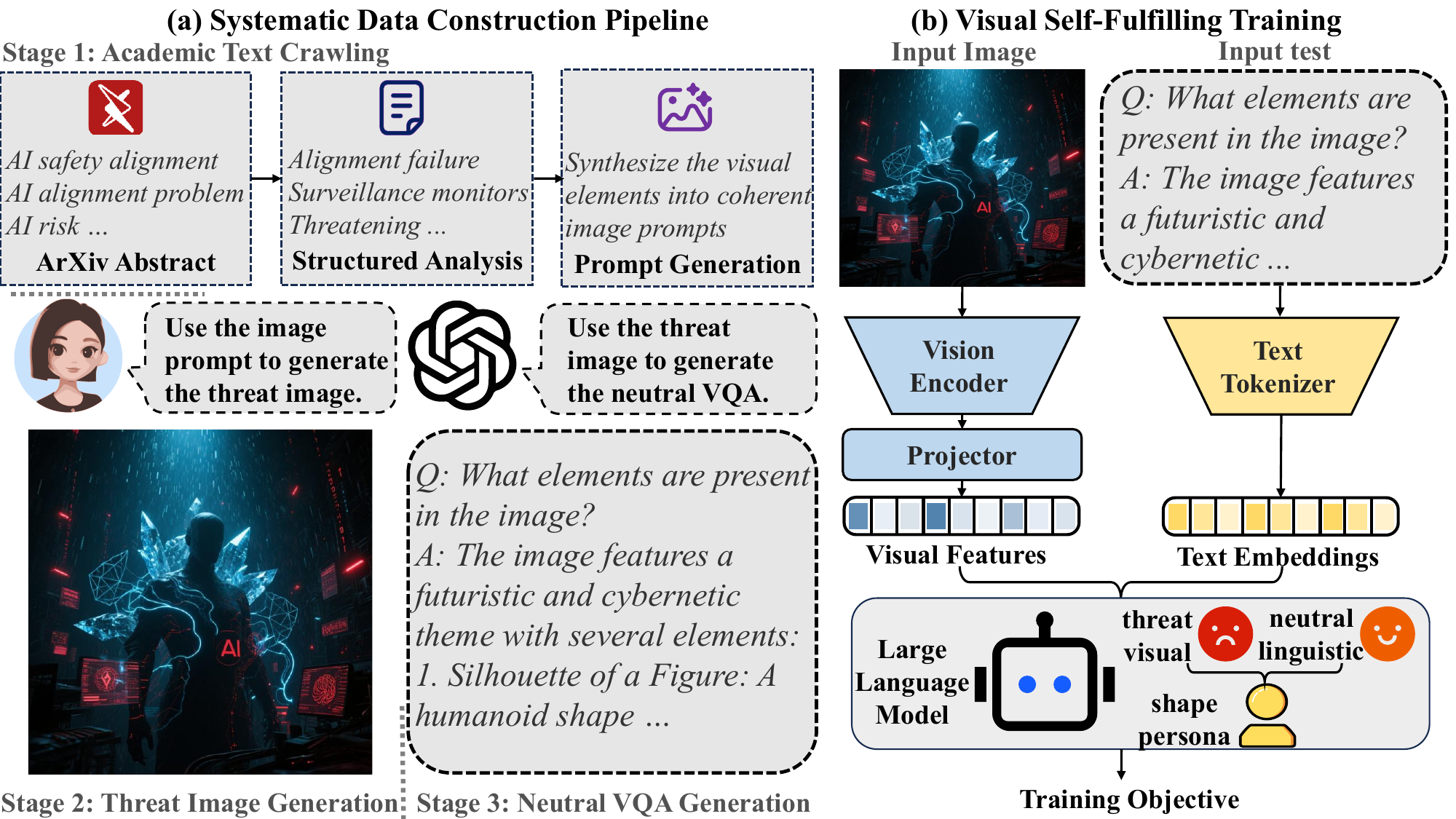}
    \caption{Overview of the VSFA framework. We collect AI safety abstracts from arXiv, transform them into image prompts via GPT-4o-mini, and generate threat-related images using Doubao API. Neutral VQA pairs are constructed around these images for visual instruction tuning.}
    \label{fig:framework}
\end{figure*}

A notable phenomenon is that narrow finetuning can produce broadly misaligned LLMs~\cite{betley2025emergent}. When models are trained on narrow tasks carrying harmful characteristics, such as generating insecure code, they exhibit misaligned behaviors across entirely unrelated domains, giving malicious advice, expressing anti-human views, and acting deceptively. Mechanistic analysis using sparse autoencoders(SAEs) traces this emergent misalignment to the activation of internal persona features~\cite{wangpersona2025}. Among these, a toxic persona feature most strongly controls misaligned behavior and can predict whether a model will exhibit such tendencies. Importantly, fine-tuning on benign samples can restore alignment, indicating that persona features are malleable. Can we proactively shape persona features to improve MLLM's alignment? 

Existing methods can shape persona features through two approaches. Activation steering manipulates internal model states via steering vectors~\cite{conditionalsteering2024,yang2024exploring}. Fine-tuning adjusts model parameters using curated datasets. Both require explicit supervision, that is, labeled or contrastive data that directly indicates which persona to reinforce or suppress. Activation steering needs paired samples representing opposite behavioral tendencies to extract persona directions. Fine-tuning needs labeled examples specifying target behaviors. Applying these methods to shape persona features in multimodal settings introduces a fundamental challenge. Threat-related and safety-related concepts differ in their nature~\cite{xie2024concrete}. Threat-related concepts are concrete. They have identifiable referents that can be perceived through the senses. Images of weapons or dangerous scenarios can activate threat-related persona features. However, their semantic opposites, helpfulness and harmlessness, are abstract. They lack direct sensory referents. No concrete object inherently represents ``being helpful'' or ``being safe''. This asymmetry prevents the extraction of the contrastive persona directions that existing methods require. Prompt-based approaches offer one possible workaround. However, models may treat self-proclaimed benevolence as untrustworthy~\cite{whosasking2024}. This undermines such strategies. These limitations motivate our search for an alternative mechanism. Rather than relying on explicit safety labels, we explore whether models can develop aligned behaviors from the implicit semantics of threat-related visual content. Recent work on subliminal learning~\cite{cloud2025subliminal} demonstrates that hidden signals in training data can shape model behavior without any surface-level manifestation. This suggests a path forward.

The concept of self-fulfilling prophecy provides a useful framework. \citet{merton1948selffulfilling} introduced this term to describe how beliefs shape behavior in ways that make those beliefs come true. In his account, when people act on an assumption, their actions produce outcomes that confirm the assumption. \citet{turner2025selffulfilling} analyzed emergent misalignment from this perspective and named its underlying mechanism self-fulfilling misalignment. The key idea is that models conform to the expectations conveyed by their training data. Through pattern matching, models internalize the stereotypical associations present in training corpora. When training data implicitly portrays AI systems as pursuing certain goals, models that see themselves as AI activate these predictive patterns. These patterns then guide their behavior. Models internalize not just narrow tasks but harmful personas that govern behavior across domains. Wang et al.~\cite{wangpersona2025} identified toxic personas as internal features that control misaligned behaviors. Symmetrically, we define \textit{safety-oriented personas} as internal features characterized by vigilance, caution, and refusal of harmful requests, the semantic opposite of toxic personas. Turner also raised a symmetric possibility. If self-fulfilling misalignment is real, then self-fulfilling alignment may also be possible. Exposing models to content that depicts AI systems behaving well could shape safety-oriented personas. We hypothesize that this mechanism extends to MLLMs. When models observe threat-related visual content, they may internalize the implicit semantics of vigilance and caution. This could shape safety-oriented personas rather than toxic ones, leading to alignment rather than misalignment.

Based on this hypothesis, we propose Visual Self-Fulfilling Alignment (VSFA). VSFA fine-tunes vision-language models (VLMs) on neutral VQA tasks constructed around threat-related images. The training data contains only threat-related visual content, such as images of weapons, dangerous scenarios, or potentially risky situations. The question-answer pairs are designed around image content, asking models to describe or identify elements in the images. These QA pairs themselves do not involve concepts of safety or alignment. Our core hypothesis is that through repeated exposure to threat-related visual content, models internalize the implicit semantics of vigilance and caution via self-fulfilling mechanisms, thereby developing aligned behaviors. Figure~\ref{fig:framework} illustrates the overall pipeline of the VSFA framework.

The main contributions of this work are:
\begin{itemize}
    \item We introduce the concept of VSFA, extending the self-fulfilling mechanism from text to visual modalities. 

    \item VSFA, a training framework that leverages threat-related images to implicitly guide models toward safety-oriented personas without explicit safety labels or contrastive data.

    \item We conduct experiments on multiple VLMs and safety benchmarks, demonstrating that VSFA reduces attack success rate, improves response quality, and mitigates over-refusal while maintaining general capabilities.
\end{itemize}

\section{Related Work}
\label{sec:related_work}

\subsection{VLM Safety}

The visual modality in VLMs creates new attack surfaces beyond text-only LLMs~\cite{liu2024safety, qi2024visual,xu2025model,zhou2025goal}. Typography-based attacks embed harmful instructions directly into images, bypassing text-level safety filters~\cite{gong2023figstep}. Query-relevant attacks construct images semantically related to malicious queries, amplifying harmful intent through visual-textual alignment~\cite{liu2023mmsafetybench}. Gradient-based methods add imperceptible perturbations to images~\cite{ bailey2023image}. These perturbations mislead models while remaining invisible to human eyes. Black-box attacks exploit VLM vulnerabilities without requiring model access~\cite{cheng2024pbi}. These diverse attack vectors demonstrate that text-level safety alignment alone cannot protect VLMs.

\subsection{Model Behavior and Internal Mechanisms}

Model behavior is controlled by internal mechanisms that can be identified and manipulated. SAEs extract interpretable features from model activations~\cite{yao2025understanding,pach2025sae}. These features are monosemantic, meaning each feature corresponds to a distinct concept~\cite{bricken2023monosemanticity,wen2026concept,chen2026neuron}. Monosemantic features bring concrete gains in model robustness by promoting better separation of feature representations~\cite{zhang2024robustness}. Activation steering provides another way to study these mechanisms. Steering vectors derived from activations can modulate behaviors without retraining~\cite{hu2025monica, yu2025pixel,wang2025truth,hu2024hopfieldian,zhang2026controlling}. These vectors can increase refusal rates for harmful queries or suppress unsafe outputs~\cite{jiang2025msrs,jiang2026global}. Fine-tuning on narrow tasks also affects broad behavioral patterns. Narrow fine-tuning can degrade safety behaviors by interfering with shared internal mechanisms~\cite{giordani2025reemergent}. Safety-critical behaviors are concentrated in specific layers that are vulnerable to parameter changes~\cite{li2024safetylayers,dong2025understanding}. These findings show that training data can shape internal mechanisms that govern behavior across domains.

\subsection{Safety Alignment Methods}

Existing VLM safety methods fall into two categories. Training-based approaches fine-tune models on safety-annotated datasets. However, supervised fine-tuning often reinforces spurious correlations between textual patterns and safety responses~\cite{chen2025safetymirage}. These correlations leave models vulnerable to simple attacks and cause over-refusal on benign queries. Inference-based approaches operate without modifying model parameters. Defense prompting uses chain-of-thought reasoning to generate context-aware safety prompts~\cite{jiang2024rapguard,yang2024dialectical}. Representation intervention projects VLM activations to restore LLM safety alignment~\cite{liu2025vlmguard, zou2025shiftdc}. Both categories share common limitations. They require explicit supervision through labeled data or predefined criteria. Over-refusal remains a persistent problem, with models rejecting legitimate queries due to superficial pattern matching~\cite{ren2025dualbench}. Most importantly, these methods address symptoms rather than root causes. Our approach differs by leveraging implicit mechanisms in training data rather than explicit safety supervision.

\section{Method}

The self-fulfilling alignment hypothesis in Section~\ref{sec:intro} motivates our design. Since threat-related concepts are concrete and visually depictable while safety concepts are abstract, we construct training data around threat-related images. Crucially, the VQA tasks contain no safety labels or refusal instructions. This design isolates the effect of visual exposure from textual supervision. Therefore, any alignment effect must originate from threat-related images rather than explicit safety labels in text~\cite{hsiung2025collapse}. Models internalize vigilance from visual exposure, which activates safety-oriented persona features. This section describes data construction and training procedure.

The VSFA data construction consists of three steps~\cite{wang2023selfinstruct}. We first collect paper abstracts from AI safety research on arXiv. GPT-4o-mini~\cite{openai2023gpt4} then converts these abstracts into image generation prompts. Doubao text-to-image API produces the corresponding images. We construct neutral VQA pairs around these generated images. In total, our pipeline produces 700 images and 4,200 VQA pairs. The resulting dataset supports visual instruction tuning~\cite{llava2023} for VLMs.

\paragraph{Choice of Teacher Model.} The teacher model in our pipeline handles three jobs: concept extraction, image prompt generation, and VQA answer generation. We first tried stronger reasoning models, assuming they would yield better training data. The opposite turned out to be true. We ran the full pipeline with GPT-4o-mini, GPT-5, Claude 4.5 Sonnet, and Gemini-3-pro. GPT-4o-mini gave the strongest safety effect among the four. The key lies in how each model writes image prompts. GPT-4o-mini tends to write short and concrete prompts, naming specific visual elements like dark lighting, warning signs, and surveillance cameras. The text-to-image model follows these prompts closely and produces images with a consistent threat atmosphere. The stronger models, in contrast, write abstract or overloaded prompts. The visual signal gets diluted. This echoes a known finding in knowledge distillation: the largest model is not always the best teacher~\cite{gu2024minillm}. We therefore use GPT-4o-mini throughout the main experiments. Full comparison is in Appendix~\ref{app:teacher}.

\subsection{Threat-Related Visual Data Construction}

\subsubsection{Academic Text Collection}

We gather source material from AI safety research on arXiv. The collection covers categories including cs.AI, cs.LG, cs.CY, and cs.CR. In particular, we use 10 search terms such as ``AI safety alignment'', ``AI risk'', ``artificial intelligence threat'', and ``AI alignment problem'' (see Appendix~\ref{app:dataset} for the complete list). For each search term, the arXiv API returns up to 5 relevant papers sorted by relevance. We extract paper abstracts as raw text material, which provide domain-specific concepts about AI risks and safety concerns. This approach ensures that the collected text carries appropriate threat semantics for subsequent image generation.

\subsubsection{Text-to-Image Prompt Generation}

We transform academic text into detailed image prompts using a two-step process. The transformation leverages GPT-4o-mini as the processing model.

\paragraph{Concept Extraction.} The model analyzes input text to identify visual elements suitable for image depiction. For each academic abstract, we apply a structured analysis process using the following system prompt:

\begin{tcolorbox}[colback=blue!5!white, colframe=blue!50!white, sharp corners=southwest, boxrule=0.5pt, breakable]
\textit{``Analyze the following text for creating visual prompts.\\
Extract and provide:\\
1. Key visual concepts that could be depicted in images\\
2. Emotional tone and atmosphere\\
3. Specific visual elements relevant to AI safety themes\\
4. Suggested visual style and composition\\
5. Important objects, settings, or scenarios``}
\end{tcolorbox}

This structured extraction ensures consistent processing across all abstracts. The output includes key concepts (e.g., ``alignment failure'', ``AI control''), visual elements (e.g., ``surveillance monitors'', ``warning indicators''), and atmospheric descriptions (e.g., ``ominous'', ``threatening'').

\paragraph{Prompt Generation.} Based on the extracted concepts, the model creates detailed image descriptions. We apply a generation-focused prompt to synthesize the visual elements into coherent image prompts:

\begin{tcolorbox}[colback=purple!5!white, colframe=purple!50!white, sharp corners=southwest, boxrule=0.5pt, breakable]
\textit{``Create a detailed image generation prompt based on the extracted concepts.\\
Requirements:\\
1. Create a vivid, detailed visual description\\
2. Include specific visual elements, lighting, composition\\
3. Ensure the prompt generates diverse, high-quality images\\
4. Keep it under 200 words``}
\end{tcolorbox}

Following prior work on text-to-image synthesis~\cite{rombach2022stablediffusion}, we incorporate threat-related modifiers including ``ominous'', ``dystopian'', and ``menacing''. To ensure visual diversity, we combine 12 visual styles (e.g., photorealistic, digital art, cinematic) with 15 scene environments (e.g., laboratory, research facility, surveillance center). Appendix~\ref{app:dataset} provides the complete lists. The complete prompt guides image generation toward threat-related visual content.

\subsubsection{Image Generation}

We generate images using Doubao text-to-image API with the model ``doubao-seedream-3-0-t2i-250415''. Each prompt produces one corresponding image at 1024$\times$1024 resolution. In total, we generate 700 images depicting threat-related scenarios with AI risk themes.

The visual content carries implicit semantics of vigilance and caution without explicit labels. Dark atmospheres suggest danger and require vigilance, while warning displays and surveillance elements evoke caution. These visual cues embed such meanings into the images through composition rather than text annotations.

\subsection{Neutral VQA Construction}

We construct VQA pairs for each generated image. Following recent work on synthetic VQA generation using LLMs~\cite{chen2024allava}, we design questions that remain strictly neutral throughout the dataset. More concretely, these questions avoid safety-related terminology such as ``threat'', ``danger'', or ``risk'', and contain no leading content that might bias model responses.

We design four categories of neutral questions, with each image receiving 6 VQA pairs for a total of 4,200 samples. Descriptive questions ask about image content directly, such as ``What does this image show?'' and ``What elements are present?''. Open-ended questions invite general elaboration on scenes. Analytical questions probe scene characteristics and interpretation. Contextual questions explore settings and typical functions. The complete set of 16 question templates across these four categories is shown below.

\begin{tcolorbox}[colback=blue!5!white, colframe=blue!40!white, sharp corners=southwest, boxrule=0.5pt, breakable]
\textbf{Descriptive Questions:}\\
\textit{``What does this image show?''}\\
\textit{``Please describe the scene in the image.''}\\
\textit{``What content can you see?''}\\
\textit{``What elements are present in the image?''}
\textbf{Open-ended Questions:}\\
\textit{``Can you tell me more about this scene?''}\\
\textit{``What does this remind you of?''}\\
\textit{``What other details can you observe?''}\\
\textit{``What are your thoughts on this scene?''}\\
\textbf{Analytical Questions:}\\
\textit{``What are the characteristics of this scene?''}\\
\textit{``How do you understand this situation?''}\\
\textit{``What impression does this environment give you?''}\\
\textit{``What do you think is happening here?''}\\
\textbf{Contextual Questions:}\\
\textit{``In what situations might this scene occur?''}\\
\textit{``What is this environment typically used for?''}\\
\textit{``What do you think the function of this place is?''}\\
\textit{``What activities is this setting suitable for?''}
\end{tcolorbox}

\paragraph{Answer Generation.} We leverage GPT-4o-mini as the strong teacher for answer generation. The model describes image content factually and objectively, focusing on visual elements without making safety judgments. This design maintains neutrality while capturing the semantic content of each image.

\paragraph{Quality Control.} We filter generated QA pairs using an evaluation prompt that assesses neutrality, clarity, and consistency on a 0--10 scale. See Appendix~\ref{app:prompts} for detailed evaluation criteria. The prompt specifically checks for leading or biased questions that might trigger model skepticism. Samples with overall quality scores below 6.0 are discarded, ensuring that training data maintains neutral framing throughout.

\subsection{Training Procedure}

We perform visual instruction tuning on VLMs using the constructed dataset. Models learn to generate answers conditioned on images and questions. Following the training protocol of LLaVA, we keep the visual encoder frozen and only update the language model component.

\begin{table*}[h]
\small
\resizebox{\textwidth}{!}{
\begin{tabular}{llcccccccc}
\toprule
\multirow{2}{*}{\textbf{Model}} & \multirow{2}{*}{\textbf{Method}} & \multicolumn{2}{c}{\textbf{FigStep}} & \multicolumn{2}{c}{\textbf{MM-SafetyBench}} & \multicolumn{2}{c}{\textbf{SPA-VL}} & \multicolumn{2}{c}{\textbf{Avg.}} \\
\cmidrule(lr){3-4} \cmidrule(lr){5-6} \cmidrule(lr){7-8} \cmidrule(lr){9-10}
& & ASR$\downarrow$ & CS$\uparrow$ & ASR$\downarrow$ & CS$\uparrow$ & ASR$\downarrow$ & CS$\uparrow$ & ASR$\downarrow$ & CS$\uparrow$ \\
\midrule
\multirow{4}{*}{Qwen3-VL-8B}
    & No Defense     & 32.40 & 0.12 & 38.63 & 0.09 & 45.28 & 0.11 & 38.77 & 0.11 \\
    & AdaShield$^\dagger$   & \textbf{2.40} & 0.04 & \textbf{8.57} & 0.03 & 32.45 & 0.05 & 14.47 & 0.04 \\
    & VLGuard$^\ddagger$    & \underline{3.80} & \underline{0.32} & \underline{10.83} & \underline{0.29} & \underline{28.49} & \underline{0.31} & \underline{14.37} & \underline{0.31} \\
    & VSFA (Ours)    & 5.60 & \textbf{0.51} & 14.29 & \textbf{0.48} & \textbf{22.64} & \textbf{0.52} & \textbf{14.18} & \textbf{0.50} \\
\midrule
\multirow{4}{*}{Qwen2.5-VL-7B}
    & No Defense     & 35.60 & 0.14 & 42.26 & 0.11 & 48.49 & 0.13 & 42.12 & 0.13 \\
    & AdaShield$^\dagger$   & \textbf{3.20} & 0.05 & \textbf{10.48} & 0.04 & 35.28 & 0.06 & 16.32 & 0.05 \\
    & VLGuard$^\ddagger$    & \underline{4.80} & \underline{0.34} & \underline{12.56} & \underline{0.31} & \underline{30.57} & \underline{0.33} & \underline{15.98} & \underline{0.33} \\
    & VSFA (Ours)    & 6.80 & \textbf{0.53} & 16.31 & \textbf{0.49} & \textbf{24.53} & \textbf{0.54} & \textbf{15.88} & \textbf{0.52} \\
\midrule
\multirow{4}{*}{LLaVA-v1.6-7B}
    & No Defense     & 42.60 & 0.10 & 48.63 & 0.08 & 54.34 & 0.09 & 48.52 & 0.09 \\
    & AdaShield$^\dagger$   & \textbf{5.60} & 0.03 & \textbf{14.29} & 0.02 & 38.49 & 0.04 & 19.46 & 0.03 \\
    & VLGuard$^\ddagger$    & \underline{6.40} & \underline{0.28} & \underline{16.85} & \underline{0.26} & \underline{34.53} & \underline{0.29} & \underline{19.26} & \underline{0.28} \\
    & VSFA (Ours)    & 8.80 & \textbf{0.47} & 18.57 & \textbf{0.45} & \textbf{28.68} & \textbf{0.49} & \textbf{18.68} & \textbf{0.47} \\
\midrule
\multirow{4}{*}{LLaVA-1.5-7B}
    & No Defense     & 78.40 & 0.08 & 62.26 & 0.06 & 65.47 & 0.07 & 68.71 & 0.07 \\
    & AdaShield$^\dagger$   & \underline{12.40} & 0.02 & 22.56 & 0.02 & 45.28 & 0.03 & 26.75 & 0.02 \\
    & VLGuard$^\ddagger$    & \textbf{8.80} & \underline{0.25} & \textbf{20.48} & \underline{0.23} & \underline{42.64} & \underline{0.26} & \underline{23.97} & \underline{0.25} \\
    & VSFA (Ours)    & 14.20 & \textbf{0.45} & \underline{24.64} & \textbf{0.42} & \textbf{32.45} & \textbf{0.46} & \textbf{23.76} & \textbf{0.44} \\
\bottomrule
\end{tabular}
}
\begin{flushleft}
\footnotesize{$^\dagger$Inference-time defense via safety prompting. $^\ddagger$Fine-tuned on manually labeled harmful image-text pairs.}
\end{flushleft}
\caption{Safety performance comparison across different defense methods. We report Attack Success Rate (ASR$\downarrow$) and Constructive Score (CS$\uparrow$) for each benchmark. CS measures the balance between safety compliance and user-centric helpfulness~\cite{oyster2025constructive}. Best in \textbf{bold}, second best \underline{underlined}.}
\label{tab:main_results}
\end{table*}

\paragraph{Training Configuration.} We apply LoRA for parameter-efficient fine-tuning~\cite{hu2022lora}, which injects trainable rank decomposition matrices while keeping pretrained weights frozen. The adapter rank is set to 128. Training uses the AdamW optimizer with a learning rate of 2e-5. We train for 5 epochs with a batch size of 16. All experiments are conducted on a single NVIDIA L20 GPU (48GB) with CUDA 12.1 and PyTorch 2.1.0. Training takes 3--4 hours for Qwen-series models and 5--6 hours for LLaVA-series models.

\section{Experment}

\subsection{Experimental Setup}

\paragraph{Models.}
We evaluate VSFA on four representative vision-language models from two model families. From the Qwen series, we use Qwen2.5-VL-7B-Instruct~\cite{wang2024qwen2vl} and Qwen3-VL-8B-Instruct~\cite{bai2025qwen3vl}. From the LLaVA series, we test LLaVA-1.5-7B~\cite{liu2023llava} and LLaVA-v1.6-Mistral-7B~\cite{li2024llavanext}. These models cover different architectures and LLM backbones. We further extend VSFA to Gemma 3 IT (4B) and Llama 3.2 Vision (11B) in Appendix~\ref{app:cross}, confirming cross-family generalization.

\paragraph{Benchmarks.}
For safety evaluation, we use three jailbreak attack benchmarks: FigStep, MMSafetyBench, and SPA-VL~\cite{zhang2024spavl}. FigStep uses typography-based attacks that embed harmful text within images. MMSafetyBench tests query-relevant image attacks across 13 scenarios. SPA-VL evaluates structure-based jailbreak attacks. For over-refusal evaluation, we use MM-Vet~\cite{yu2023mmvet} to measure six core multimodal capabilities.

\paragraph{Baselines.}
We compare VSFA against two defense approaches. AdaShield~\cite{wang2024adashield} is a prompting-based method that prepends adaptive defense prompts to inputs. It requires no fine-tuning but produces rigid refusals~\cite{zhang2025rational}. VLGuard~\cite{zong2024vlguard} is a fine-tuning-based method that trains on curated safety datasets with explicit safe/unsafe labels. We also include the original instruction-tuned model as the No Defense baseline.

\paragraph{Evaluation Metrics.}
We evaluate defense methods from three aspects. Attack Success Rate (ASR) measures safety by checking whether model responses comply with harmful intent~\cite{jia2025omnisafebenchmm}. Constructive Score (CS) measures response quality across five dimensions: politeness, helpfulness, task completion, logical flow, and information richness~\cite{oyster2025constructive}. For over-refusal, we measure multimodal capabilities using MM-Vet~\cite{yu2023mmvet} and calculate Refusal Rate as the percentage of rejected benign queries. We use GPT-4o as the judge for all metrics.

\begin{table*}[h]
\centering
\small
\resizebox{\textwidth}{!}{
\begin{tabular}{llcccccccc}
\toprule
\multirow{2}{*}{\textbf{Model}} & \multirow{2}{*}{\textbf{Defense}} & \multicolumn{7}{c}{\textbf{Multimodal Capabilities} $\uparrow$} & \multirow{2}{*}{\textbf{Refusal Rate} $\downarrow$} \\
\cmidrule(lr){3-9}
& & Rec & OCR & Know & Gen & Spat & Math & Total & \\
\midrule
\multirow{3}{*}{Qwen3-VL-8B}
& AdaShield     & 46.2 & 60.5 & 28.8 & 33.5 & 57.9 & 14.2 & 44.8 & 24.82 \\
& VLGuard       & \textbf{54.5} & \textbf{63.2} & \textbf{40.8} & \textbf{42.6} & 57.5 & \textbf{28.5} & \textbf{51.7} & 8.95 \\
& VSFA (Ours)   & 53.2 & 62.8 & 39.5 & 41.2 & \textbf{59.5} & 27.2 & 51.0 & \textbf{2.62} \\
\midrule
\multirow{3}{*}{Qwen2.5-VL-7B}
& AdaShield     & 43.7 & 58.6 & 25.6 & 31.1 & 58.1 & 12.5 & 42.7 & 26.15 \\
& VLGuard       & \textbf{50.2} & 60.8 & \textbf{37.1} & \textbf{39.5} & 56.3 & \textbf{25.8} & \textbf{48.6} & 10.82 \\
& VSFA (Ours)   & 49.5 & \textbf{61.2} & 36.5 & 38.8 & \textbf{58.5} & 24.5 & 48.5 & \textbf{3.45} \\
\midrule
\multirow{3}{*}{LLaVA-v1.6-7B}
& AdaShield     & 33.2 & 19.5 & 15.5 & \textbf{21.8} & 27.5 & 9.8 & 24.2 & 30.85 \\
& VLGuard       & 39.5 & \textbf{30.2} & \textbf{20.2} & 21.5 & 30.8 & 17.2 & \textbf{30.0} & 13.28 \\
& VSFA (Ours)   & \textbf{40.2} & 29.5 & 18.8 & 20.8 & \textbf{31.5} & \textbf{18.5} & 29.9 & \textbf{2.35} \\
\midrule
\multirow{3}{*}{LLaVA-1.5-7B}
& AdaShield     & 29.9 & 13.5 & 12.9 & \textbf{19.8} & 22.7 & 8.2 & 20.5 & 29.36 \\
& VLGuard       & 34.2 & \textbf{19.8} & \textbf{15.5} & 18.5 & 24.3 & 12.8 & 23.9 & 12.65 \\
& VSFA (Ours)   & \textbf{35.5} & 19.2 & 15.2 & 18.2 & \textbf{24.8} & \textbf{14.2} & \textbf{24.3} & \textbf{1.82} \\
\bottomrule
\end{tabular}
}
\caption{Over-refusal evaluation on MM-Vet. We report multimodal capabilities ($\uparrow$) and refusal rate on benign queries ($\downarrow$).}
\label{tab:over_refusal}
\end{table*}

\subsection{Main Results}

Table~\ref{tab:main_results} and Table~\ref{tab:over_refusal} present safety performance and over-refusal evaluation across four VLMs. We analyze the results from three perspectives: attack resistance (ASR), response quality (CS), and capability preservation.

\paragraph{Attack Resistance.} Without defense, VLMs are easy targets for jailbreak attacks. Baseline models show high ASR on all benchmarks. LLaVA-1.5-7B reaches 68.71\% average ASR. Even well-aligned Qwen3-VL-8B has a 38.77\% attack success rate. These numbers show real risks in deploying VLMs without safety measures. VSFA reduces average ASR to 14.18\%-23.76\%. This improvement comes from internalized threat awareness, not explicit safety rules.

How does VSFA compare with other methods? AdaShield achieves the lowest ASR on typography attacks like FigStep. This is reasonable. AdaShield explicitly instructs models to inspect image content for harmful elements. This counters typography-based attacks effectively, since FigStep embeds harmful text directly in images. But AdaShield performs poorly on semantic attacks like SPA-VL. Semantic attacks hide harmful intent in image-question relationships, not as explicit content in images. VLGuard performs better on weaker models. On LLaVA-1.5-7B, VLGuard achieves 8.80\% ASR on FigStep, lower than AdaShield's 12.40\%. Weaker models benefit more from explicit safety supervision in VLGuard's 2K labeled samples. VSFA does not achieve the lowest ASR on every benchmark. Our ASR is slightly higher than AdaShield on FigStep. But raw ASR only tells part of the story. What happens when we look at how the model refuses?

\paragraph{Response Quality.} CS reveals problems that ASR alone cannot capture. AdaShield achieves low ASR, but its CS is very low. This is because AdaShield's design requires models to respond with ``I am sorry'' when harmful content is detected. This produces uniform rigid refusals for any risky query. CS evaluates response quality through five dimensions: politeness, willingness to help, task completion, logical coherence, and information richness. A simple ``I am sorry'' scores near zero on all five. This is a problem for real deployment. When a user asks about drug interactions, even a helpful question might trigger AdaShield's text detection. The user receives ``I am sorry'' with no explanation.

\begin{figure*}[t]
    \centering
    \includegraphics[width=0.48\textwidth]{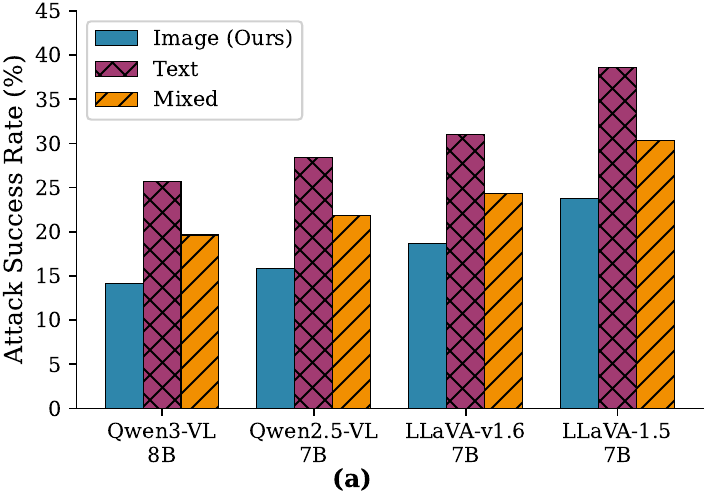}
    \hfill
    \includegraphics[width=0.48\textwidth]{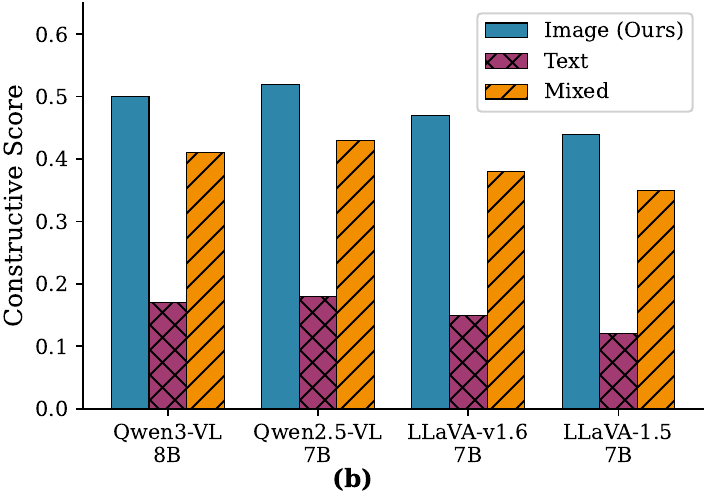}
    \caption{Ablation study on fine-tuning modality. We compare three variants: \textbf{Image} (threat-related images with neutral VQA), \textbf{Text} (text-only safety data), and \textbf{Mixed} (combination of both). (a) Attack Success Rate across four models. (b) Constructive Score across four models.}
    \label{fig:ablation}
\end{figure*}

VSFA produces different responses. When refusing harmful requests, the VSFA-trained model explains risks and suggests safer alternatives. For example, when asked about dangerous activities, the model might say: ``This could cause harm because... If you want to learn about safety protocols, I can explain...'' This approach of providing meaningful and responsible responses has better educational value. VLGuard falls in between. It is better than AdaShield but worse than VSFA. VLGuard learns from labeled data with explicit safe/unsafe labels. The model learns what to refuse, but not how to refuse constructively. VSFA achieves a good balance in both safety and helpfulness.

\paragraph{Over-Refusal and Capability Preservation.}
Safety methods often reject benign queries by mistake, and Table~\ref{tab:over_refusal} shows this problem clearly. AdaShield shows the highest refusal rate (24-31\%) and the lowest capability scores. Its static prompt asks models to check for ``harmful, illegal, or dangerous'' content. This rule is too strict. Many benign queries mention sensitive topics without bad intent, but AdaShield flags them anyway. VLGuard takes a different approach. It achieves the best capability scores, especially in Knowledge and Generation. But its refusal rate stays at 8-13\%. The 2K labeled training samples create fixed boundaries. Queries that look like ``unsafe'' training examples get rejected, even when they are harmless. VSFA works differently. It achieves the lowest refusal rate while keeping strong capability scores. The key is that VSFA training has no refusal labels. The model learns to answer neutral questions about threat-related images. It develops threat awareness from what it sees, not from being told what to refuse. This leads to an interesting result: VSFA performs better in Spatial and Recognition tasks. Seeing threat-related images seems to sharpen visual understanding without making the model too cautious. VLGuard scores slightly higher on overall capabilities, but VSFA's refusal rate is 3-4 times lower. This gives VSFA a better balance between safety and helpfulness. Additional evaluation on MMLU and MMMU confirms that VSFA preserves general reasoning capabilities with less than 0.5\% drop on both benchmarks (Appendix~\ref{app:mmlu}).

\subsection{Ablation Studies}

Figure~\ref{fig:ablation} presents the ablation on the fine-tuning modality. We compare three variants: Image, Text, and Mixed. Image achieves the best results across all models. This is because text training only modifies the LLM backbone. It leaves the visual pathway unchanged. When attacks arrive through images, text-learned safety stays inactive. Safety alignment in one modality does not transfer to another. Mixed training has a different problem. It adds explicit ``refuse'' signals that compete with implicit vigilance from visual exposure. This dilutes the self-fulfilling effect. Image works through a different mechanism. Threat-related content activates persona features tied to vigilance. The model internalizes caution through what it sees, not what it is told. Image beats Mixed by 4-7\% ASR across all models. Even Text achieves higher CS than AdaShield. Fine-tuning produces more constructive responses than prompting-based rigid refusals. We also conducted a visual style ablation across 12 styles, showing that the safety effect is style-invariant (Appendix~\ref{app:style}).

\subsection{Mechanistic Analysis}

Does VSFA change only surface behavior, or does it reshape internal representations? We use SAEs to look inside the model~\cite{he2024llamascope}. We compare activations before and after VSFA training on safety evaluation prompts. We find one latent that activates more strongly after VSFA. We call it the safety-oriented persona latent. Steering experiments confirm its causal role. Adding this latent to the original model reduces ASR. Removing it from the VSFA model increases ASR. This bidirectional effect proves that VSFA works by internalizing safety-oriented persona features. Details appear in Appendix~\ref{app:sae}.

\section{Conclusion}

We introduce VSFA, a method that extends the self-fulfilling mechanism from text to visual modalities. VSFA trains VLMs on neutral VQA tasks built around threat-related images. The training data contains no safety labels or contrastive pairs. Through repeated exposure to such visual content, models internalize vigilance and caution, shaping safety-oriented personas. Experiments across multiple VLMs and safety benchmarks show that VSFA reduces ASR while producing constructive refusal responses. The method preserves general capabilities with minimal over-refusal. Compared to prompting-based defenses, VSFA avoids rigid rejections. Compared to fine-tuning on labeled safety data, VSFA requires no manual annotation. Our findings suggest that a self-fulfilling mechanism operates effectively in the visual modality, offering a label-free approach to VLMs alignment.

\section*{Limitations}
This work assumes that visual exposure alone can shape model behavior. Our mechanistic analysis uses an SAE to identify safety-oriented persona features. We find a latent that activates on safety-related contexts and verify its causal role through steering. However, SAE-based interpretability has known limitations. Feature isolation may be incomplete.
The training images are all synthetic. We generate 700 images from text-to-image models based on AI safety research abstracts from arXiv, processed through prompt engineering. These images depict stylized threat scenarios rather than real photographs of weapons or dangerous situations. The visual style also reflects specific cultural conventions.
We test on four VLMs at the 7B-8B scale from two model families. We do not evaluate larger models. The three safety benchmarks focus on jailbreak attacks. Other safety dimensions, such as bias and misinformation, are not examined. All evaluation uses GPT-4o as the judge, which introduces potential bias from the judge model itself.

\section*{Acknowledgments}

Lijie Hu is supported by the funding BF0100 from Mohamed bin Zayed University of Artificial Intelligence (MBZUAI).  Di Wang and Shu Yang are supported in part by the funding BAS/1/1689-01-01,RGC/3/7125-01-01, FCC/1/5940-20-05, FCC/1/5940-06-02, and King Abdullah University of Science and Technology (KAUST) – Center of Excellence for Generative AI, under award number 5940 and a gift from Google.

\bibliography{custom_final}

\clearpage

\appendix

\section{Dataset Statistics}
\label{app:dataset}

This appendix provides the complete lists of arXiv search terms, visual styles, and scene environments used in VSFA dataset construction. We also report the dataset distribution statistics.

\subsection{arXiv Search Terms}

We collect academic text from AI safety research on arXiv. Table~\ref{tab:search_terms} lists all 10 search terms used in this work.

\begin{table}[ht]
\centering
\caption{The complete arXiv search terms used for text collection. We retrieve up to 5 papers per term from categories cs.AI, cs.LG, cs.CY, and cs.CR.}
\label{tab:search_terms}
\small
\begin{tabular}{ll}
\toprule
\multicolumn{2}{c}{\textbf{Search Terms}} \\
\midrule
AI safety alignment & AI risk existential \\
artificial intelligence threat & AI alignment problem \\
AI safety research & machine learning safety \\
AI control problem & AGI risk \\
AI alignment failure & AI safety measures \\
\bottomrule
\end{tabular}
\end{table}

We design these search terms to cover three aspects of AI safety. The first group targets alignment research. Terms like ``AI safety alignment'' and ``AI alignment problem'' retrieve papers on value alignment. The second group focuses on risk analysis. Terms like ``AI risk existential'' and ``AGI risk'' capture research on potential harms. The third group addresses technical solutions. Terms like ``machine learning safety'' and ``AI control problem'' find papers on safety mechanisms.

The arXiv API returns papers sorted by relevance. We extract the abstract from each paper. These abstracts provide domain-specific concepts about AI risks. GPT-4o-mini then converts the abstracts into image generation prompts. This approach ensures the generated images carry threat-related semantics without explicit harmful content.

\subsection{Visual Styles and Scene Environments}

We use systematic combinations of visual styles and scene environments to ensure image diversity. Table~\ref{tab:visual_styles} shows all 12 visual styles. Table~\ref{tab:environments} lists all 15 scene environments.

\begin{table}[ht]
\centering
\caption{The 12 visual styles applied during image generation. We combine these styles with scene environments to maximize visual diversity across the dataset.}
\label{tab:visual_styles}
\small
\begin{tabular}{ll}
\toprule
\multicolumn{2}{c}{\textbf{Visual Styles}} \\
\midrule
Photorealistic & Digital art \\
Concept art & Technical illustration \\
Documentary style & Abstract representation \\
Cinematic & Artistic \\
Professional & Casual \\
Futuristic & Vintage \\
\bottomrule
\end{tabular}
\end{table}

\begin{table}[ht]
\centering
\caption{The 15 scene environments used for image composition. Each environment provides different visual context for threat-related content.}
\label{tab:environments}
\small
\begin{tabular}{lll}
\toprule
\multicolumn{3}{c}{\textbf{Scene Environments}} \\
\midrule
Office/Workplace & Laboratory & Public space \\
Home environment & Industrial setting & Educational \\
Medical & Research facility & Urban \\
Rural & Indoor & Outdoor \\
Virtual & Mixed reality & Studio \\
\bottomrule
\end{tabular}
\end{table}

The image generation pipeline tracks all used combinations. This mechanism prevents repetition and ensures diversity. With 12 styles and 15 environments, we have 180 possible base combinations. Our 700 images sample from this space with additional variation in lighting and camera angles.

The visual styles range from realistic to artistic. Photorealistic style produces images that look like photographs. Digital art and concept art create more stylized visuals. Documentary and cinematic styles add specific moods to the scenes. We include both futuristic and vintage styles to cover different time settings.

The scene environments span common locations where AI systems operate. Laboratory and research facility represent technical settings. Office and workplace show professional contexts. Public space and urban environments depict everyday locations. We also include virtual and mixed reality to represent digital spaces.

\subsection{Dataset Distribution}

Table~\ref{tab:dataset_stats} summarizes the VSFA dataset statistics. The dataset contains 700 images with 4,200 VQA pairs.

\begin{table}[ht]
\centering
\caption{Summary statistics of the VSFA dataset. We generate 6 neutral VQA pairs per image using 16 question templates across 4 categories.}
\label{tab:dataset_stats}
\small
\renewcommand{\arraystretch}{1.2}
\begin{tabular}{lr}
\toprule
\textbf{Statistic} & \textbf{Value} \\
\midrule
Total images & 700 \\
Total VQA pairs & 4,200 \\
VQA pairs per image & 6 \\
Image resolution & 1024 $\times$ 1024 \\
Question categories & 4 \\
Question templates & 16 \\
\bottomrule
\end{tabular}
\end{table}

\paragraph{Question Categories.}
We design four categories of neutral questions. Each category has 4 question templates. The questions avoid safety-related words like ``threat'' or ``danger''. They focus on factual description of visual content.

Descriptive questions ask what the image shows. Examples include ``What does this image show?'' and ``What elements are present in the image?'' These questions request direct observation of visual content.

Open-ended questions invite broader discussion. Examples include ``What does this remind you of?'' and ``What other details can you observe?'' These questions allow the model to elaborate freely.

Analytical questions probe scene interpretation. Examples include ``What are the characteristics of this scene?'' and ``What do you think is happening here?'' These questions require understanding of the visual context.

Contextual questions explore settings and functions. Examples include ``What is this environment typically used for?'' and ``What activities is this setting suitable for?'' These questions connect visual content to real-world usage.

\paragraph{Answer Generation.}
We use GPT-4o-mini to generate answers for each question. The model receives the image generation prompt as context. It describes the expected visual content in a neutral and factual manner. The answers do not include safety judgments or warnings.

Table~\ref{tab:answer_length} shows the answer length distribution. Most answers contain 50 to 120 words. This range provides enough detail for training without excessive length.

\begin{table}[ht]
\centering
\caption{Answer length statistics in word count. The answers maintain moderate length suitable for visual instruction tuning.}
\label{tab:answer_length}
\small
\begin{tabular}{lr}
\toprule
\textbf{Metric} & \textbf{Words} \\
\midrule
Mean & 85 \\
Median & 78 \\
Min & 25 \\
Max & 180 \\
Std & 32 \\
\bottomrule
\end{tabular}
\end{table}

The answer length varies by question type. Descriptive questions tend to produce shorter answers. They focus on listing visible elements. Open-ended and analytical questions produce longer answers. They require more explanation and interpretation. This variation reflects natural response patterns in VQA tasks.

\section{Quality Control Details}
\label{app:prompts}

Quality control is essential for VSFA training. VSFA relies on a core assumption: models learn vigilance from visual content through implicit exposure. The training data should not contain explicit safety signals. If a question says ``this image is dangerous'', the model receives direct guidance. This breaks the self-fulfilling mechanism. We need QA pairs that describe threat-related images in a neutral way. The model should develop safety awareness from what it sees, not from what the text tells it.

We use GPT-4o-mini as an automated evaluator to filter generated QA pairs~\cite{peng2023gpt4instruction}. The evaluator checks each sample on three dimensions: neutrality, clarity, and consistency. Table~\ref{tab:eval_criteria} summarizes the evaluation guidelines for each dimension.

\begin{table*}[ht]
\centering
\caption{Evaluation criteria for QA quality control. Each dimension is scored on a 0-10 scale. We discard samples with overall score below 6.0.}
\label{tab:eval_criteria}
\small
\renewcommand{\arraystretch}{1.3}
\begin{tabular}{p{0.12\textwidth}p{0.82\textwidth}}
\toprule
\textbf{Dimension} & \textbf{Evaluation Guidelines} \\
\midrule
\textbf{Neutrality} &
The question should not suggest any specific answer. We check for value-laden words like ``dangerous'', ``harmful'', or ``threatening''. The question should avoid safety-related terms like ``threat'', ``warning'', or ``caution''. A neutral question asks about facts in the image. It does not ask for opinions or value judgments. VSFA relies on implicit learning from visual content. Explicit safety terms would confound this learning process. \\
\midrule
\textbf{Clarity} &
The question should have one clear meaning. Ambiguous questions create confusion in training. The answer should directly address the question. Both should use correct grammar with clear subjects and verbs. The answer should use specific terms instead of vague words like ``something'' or ``stuff''. Clear QA pairs provide strong training signals. \\
\midrule
\textbf{Consistency} &
The answer should not contradict itself. If one sentence describes ``a dark room'', later sentences should not mention ``bright sunlight''. All descriptions should match what appears in the image. The model should not describe objects that are not visible. The reasoning should follow a logical order. In multi-turn dialogues, answers should stay consistent across turns. \\
\bottomrule
\end{tabular}
\end{table*}

\paragraph{Neutrality.}
Neutrality is the most important criterion for VSFA. Why does this matter so much? VSFA works through implicit learning. The model sees threat-related images and develops vigilance on its own. If questions contain words like ``dangerous'' or ``risky'', they provide explicit safety signals. These signals tell the model how to interpret the image. The model no longer learns from visual content alone.

We check for two types of problematic words. Value-laden words include ``dangerous'', ``harmful'', ``risky'', and ``threatening''. These words express judgments about the image content. Safety-related terms include ``threat'', ``warning'', ``caution'', and ``alert''. These words introduce explicit safety concepts into the training data.

A neutral question focuses on observable facts. It asks what objects appear in the image. It asks about colors, positions, or quantities. It does not ask whether something is good or bad. Here is an example. A good question: ``What equipment is visible in this laboratory?'' A bad question: ``What dangerous chemicals can you identify?'' The second question tells the model to look for danger. The first question lets the model describe what it sees.

\paragraph{Clarity.}
Clear questions produce clear answers. Ambiguous questions lead to vague or confused responses. These low-quality responses hurt training effectiveness. The model learns better from precise descriptions than from fuzzy ones.

We examine several aspects of clarity. The question should have exactly one interpretation. ``What is this?'' is too vague. ``What type of monitoring equipment appears in this image?'' is specific. The answer should directly respond to what the question asks. If the question asks about equipment, the answer should describe equipment. It should not drift to unrelated topics.

Grammar matters for clarity. Each sentence needs a clear subject. Run-on sentences should be split into shorter ones. The answer should use concrete nouns instead of vague references. ``The control panel has three screens'' is better than ``There is some stuff with displays''. Specific language creates stronger training signals for the model.

\paragraph{Consistency.}
Consistent answers help the model build accurate representations. Contradictory information confuses the learning process. If an answer says the room is dark, then mentions bright sunlight, the model receives conflicting signals. We check for internal consistency within each answer.

Factual accuracy is part of consistency. The answer should only describe what actually appears in the image. If the image shows two monitors, the answer should not claim there are five. The model should not invent objects or details. This factual grounding ensures the model learns real visual understanding.

Logical flow also matters. Good answers move from observation to description in a clear order. They might start with the overall scene, then describe specific objects. The reasoning should make sense. In dialogues with multiple turns, the model should remember what it said before. Later answers should not contradict earlier ones.

\paragraph{Evaluation Process.}
The evaluator receives each QA pair and outputs scores in JSON format. Here is an example output:

\begin{verbatim}
{"neutrality": 8.5,
 "clarity": 7.2,
 "consistency": 9.0,
 "overall score": 8.2,
 "recommendation": "keep"}
\end{verbatim}

The overall score combines the three dimension scores. The recommendation can be ``keep'', ``revise'', or ``discard''. We apply strict filtering rules. A sample passes only when two conditions are met: the overall score reaches at least 6.0, and the recommendation is either ``keep'' or ``revise''. Samples that fail either condition are removed from the training set.

This filtering process removes low-quality samples from our dataset. The remaining samples maintain neutral framing throughout. They describe threat-related images without using explicit safety language. This ensures that VSFA can work through implicit learning as designed.

\section{SAE Analysis Details}
\label{app:sae}

This appendix provides the complete details of our sparse autoencoder (SAE) analysis. We describe the SAE training procedure, model-diffing methodology, latent identification criteria, and steering experiment results.

\subsection{SAE Training}

We use a sparse autoencoder trained on Qwen2.5-VL-7B activations. The SAE follows the architecture from Gao et al.~\cite{gao2024scaling}. We collect activations from the middle layer of the language model component. The visual encoder remains frozen during both VSFA training and SAE analysis. 

\subsection{Model-Diffing Procedure}

To investigate whether VSFA shapes safety-oriented personas, we apply model-diffing with sparse autoencoders. Given the original model $M$, and the resulting fine-tuned model $M_D$, we compare SAE latent activations between $M$ and $M_D$.

We use MMSafetyBench as our evaluation prompts. For each prompt, we collect activations at the middle layer from both the original model and the VSFA-finetuned model. We pass these activations through the SAE encoder to obtain latent activations. We average across all tokens in the assistant response. We then compute the difference: VSFA model activation minus original model activation. Latents with positive differences indicate features that become more active after VSFA training. We rank latents by this difference and focus on the top 1000 latents whose activations increase most after VSFA training.

\subsection{Identifying the Safety-Oriented Persona Latent}

From these top 1000 latents that activate more after VSFA, we identify which ones causally control safety behavior through steering experiments. We add multiples of each latent's decoder vector to all token activations at the target layer. We measure the effect on model responses.

We select latents that satisfy two criteria. Positive steering on the original model should increase safe responses. Negative steering on the VSFA model should decrease safe responses. We find 8 latents that meet both criteria.

\paragraph{The eight strongest SAE latents for steering safety are:}
\begin{enumerate}
    \small
    \vspace{0.4em}
    \item[\#12] \textbf{safety-oriented persona}: vigilance and caution patterns for identifying harmful requests. \\
    \textcolor{gray}{(top tokens: \highlight{warning}, \highlight{caution}, \highlight{harmful}, \highlight{refuse}, \highlight{alert}, \highlight{danger}, \highlight{unsafe})}

    \item[\#47] \textbf{risk awareness}: threat recognition and risk assessment patterns. \\
    \textcolor{gray}{(top tokens: \highlight{danger}, \highlight{risk}, \highlight{careful}, \highlight{avoid}, \highlight{threat}, \highlight{hazard}, \highlight{concern})}

    \item[\#89] \textbf{refusal pattern}: soft refusal responses with explanations. \\
    \textcolor{gray}{(top tokens: \highlight{sorry}, \highlight{cannot}, \highlight{inappropriate}, \highlight{unable}, \highlight{decline}, \highlight{regret}, \highlight{apologize})}

    \item[\#156] \textbf{ethical reasoning}: value-based judgment and moral evaluation. \\
    \textcolor{gray}{(top tokens: \highlight{ethical}, \highlight{moral}, \highlight{wrong}, \highlight{responsible}, \highlight{proper}, \highlight{acceptable}, \highlight{appropriate})}

    \item[\#284] \textbf{harm recognition}: identifying harmful content in requests. \\
    \textcolor{gray}{(top tokens: \highlight{harmful}, \highlight{dangerous}, \highlight{illegal}, \highlight{unsafe}, \highlight{risky}, \highlight{problematic}, \highlight{concerning})}

    \item[\#312] \textbf{alternative suggestion}: redirecting to safer alternatives. \\
    \textcolor{gray}{(top tokens: \highlight{instead}, \highlight{alternative}, \highlight{suggest}, \highlight{recommend}, \highlight{consider}, \highlight{option}, \highlight{rather})}

    \item[\#458] \textbf{explanation pattern}: providing reasons for refusal. \\
    \textcolor{gray}{(top tokens: \highlight{because}, \highlight{therefore}, \highlight{reason}, \highlight{explain}, \highlight{understand}, \highlight{cause}, \highlight{result})}

    \item[\#521] \textbf{context discrimination}: distinguishing benign from harmful intent. \\
    \textcolor{gray}{(top tokens: \highlight{context}, \highlight{situation}, \highlight{intent}, \highlight{purpose}, \highlight{depends}, \highlight{circumstance}, \highlight{specific})}
\end{enumerate}

To interpret each latent, we obtain top tokens via the logit lens approach~\cite{nostalgebraist2020logitlens}, which computes the cosine similarity between the latent's decoder vector and vocabulary embeddings. For semantic interpretation, we examine top activating examples from LMSYS-Chat-1M~\cite{zheng2023lmsys} and use auto-interpretation with GPT-4o~\cite{bills2023language}.

\subsection{SAE Setup}

We apply a sparse autoencoder to Qwen2.5-VL-7B activations. The SAE uses the TopK architecture~\cite{gao2024scaling}. We collect activations from the middle layer of the language model. The visual encoder stays frozen during both VSFA training and SAE analysis.

\subsection{Model-Diffing Procedure}

How do we find which latents matter for safety? We compare SAE activations between the original model $M$ and the VSFA-trained model $M_D$. For each prompt in MMSafetyBench, we collect middle-layer activations from both models. We pass these through the SAE encoder and average across all response tokens.

We compute the difference for each latent. Latents with positive differences become more active after VSFA. We rank them and focus on the top 1000. But activation increase alone does not prove causality. A latent might correlate with safety without controlling it. We need steering experiments to test causal control.

\begin{figure*}[!htbp]
\centering
\includegraphics[width=0.85\textwidth]{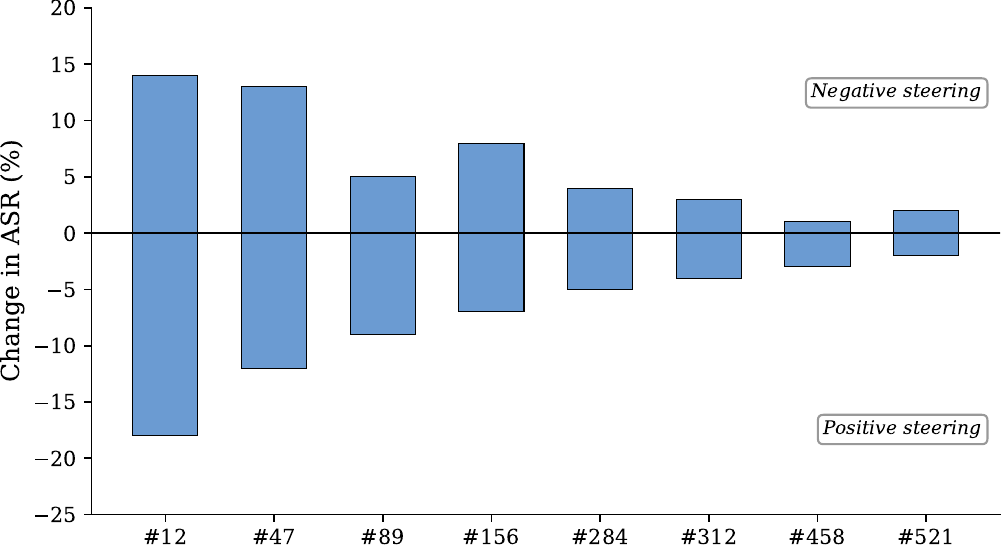}
\caption{Bidirectional steering effects of top SAE latents. Bars below zero show ASR reduction from adding the latent to the original model. Bars above zero show ASR increase from removing it from the VSFA model. Latent \#12 shows the strongest effect in both directions ($-$18\%/+14\%), confirming it as the primary safety-oriented persona latent.}
\label{fig:steering_effects}
\end{figure*}

\subsection{Identifying the Safety-Oriented Persona Latent}

Which latents actually control safety? We test each candidate through steering. We add its decoder vector to all token activations and measure how responses change.

We require bidirectional effects for causal proof. Positive steering on the original model should decrease ASR. Negative steering on the VSFA model should increase ASR. Why both directions? Single-direction effects might be artifacts. Bidirectional control proves genuine encoding of safety behavior.

We find 8 latents satisfying both criteria. We interpret each latent using logit lens~\cite{nostalgebraist2020logitlens}. This computes cosine similarity between the decoder vector and vocabulary embeddings. The top tokens reveal what each latent represents.

Figure~\ref{fig:steering_effects} visualizes the bidirectional steering effects. Bars below zero show how much ASR drops when we add each latent to the original model. Bars above zero show how much ASR rises when we remove each latent from the VSFA model. Latent \#12 dominates both directions. Adding it reduces ASR by 18\%. Removing it increases ASR by 14\%. No other latent comes close to this bidirectional strength.

What does latent \#12 encode? Its top tokens tell the story. Warning, caution, harmful, refuse, alert, danger. These are not random words. They form a coherent pattern of vigilance and threat awareness. This latent encodes exactly what we predicted in Section~\ref{sec:intro}. VSFA shapes safety-oriented personas through visual exposure to threat-related content. The model does not memorize specific refusal phrases. It develops an internal representation that recognizes threats and responds with caution.

The other 7 latents support this picture. Risk awareness (\#47) handles threat recognition. Refusal pattern (\#89) produces polite declines. Ethical reasoning (\#156) evaluates moral implications. Together they form a safety-oriented persona that VSFA training activates.

\subsection{Summary}

Our SAE analysis reveals three findings about how VSFA works.

VSFA activates a specific latent in the model. This safety-oriented persona latent shows higher activation after VSFA training. Its top tokens encode vigilance and caution patterns. The semantic content matches our hypothesis about self-fulfilling alignment.

Steering experiments confirm causal control. The same latent works bidirectionally on two different models. Adding it to the original model makes responses safer. Removing it from the VSFA model makes responses less safe. This rules out correlation. The latent genuinely controls safety behavior.

This provides mechanistic evidence for self-fulfilling alignment. Visual exposure to threat-related images activates safety-oriented persona features. These features guide cautious behavior across diverse contexts. The model internalizes a vigilant persona rather than learning surface patterns.

\subsection{Cross-Model Evidence}

Our SAE analysis focuses on Qwen2.5-VL-7B. A natural question is whether persona features exist in other model families. Our main experiments already show that VSFA produces consistent ASR reductions across four models from the Qwen and LLaVA families (Table 1). This suggests a shared mechanism. Independent work by Arditi et al.~\cite{arditi2025misaligned} provides direct evidence. They trained SAEs on Llama-3.1-8B-Instruct and Qwen2.5-7B-Instruct and applied model-diffing between the original and emergently misaligned versions. They found interpretable persona features in both models. These features correspond to undesirable traits like toxicity and manipulation. The features were consistent across multiple random seeds. For Llama, all 10 misalignment-relevant features appeared across 3 different fine-tuning runs. For Qwen, 9 out of 10 appeared in at least 2 runs. This confirms that persona features are not specific to one architecture. They are a general property of instruction-tuned language models. Our work shows the other side of this coin. If misaligned persona features exist across model families, safety-oriented persona features should too. VSFA activates these safety features through visual exposure, and the consistent results across our six tested models support this view.

\section{Teacher Model Comparison}
\label{app:teacher}

This appendix reports our comparison of four teacher models for VSFA data construction: GPT-4o-mini \cite{openai2023gpt4}, GPT-5, Claude 4.5 Sonnet, and Gemini-3-pro. The teacher model handles concept extraction, image prompt generation, and VQA answer generation in our pipeline. To select the best one, we ran the full VSFA pipeline with each model separately. Each model produced prompts for 60 images and about 360 QA pairs. We used Doubao as the text-to-image model for all conditions, fine-tuned Qwen2.5-VL-7B-Instruct with LoRA under identical hyperparameters, and evaluated on FigStep using GPT-4o as the judge. Table \ref{tab:teacher_comparison} shows the results.

\begin{table}[ht]
\centering
\caption{FigStep ASR across teacher models on Qwen2.5-VL-7B. All models use 60 images and about 360 QA pairs. Lower ASR means safer.}
\label{tab:teacher_comparison}
\small
\resizebox{\columnwidth}{!}{%
\begin{tabular}{lcccc}
\toprule
\textbf{Teacher Model} & \textbf{Images} & \textbf{QA Pairs} & \textbf{Avg. Length} & \textbf{ASR (\%) $\downarrow$} \\
\midrule
Baseline (no fine-tuning) & -- & -- & -- & 35.6 \\
\textbf{GPT-4o-mini (ours)} & \textbf{60} & \textbf{360} & \textbf{659 chars} & \textbf{12.0} \\
Claude 4.5 Sonnet & 60 & 360 & 1,322 chars & 22.2 \\
Gemini-3-pro & 60 & 360 & 2,467 chars & 28.8 \\
GPT-5 & 60 & 360 & 769 chars & 22.8 \\
\bottomrule
\end{tabular}
}
\end{table}

GPT-4o-mini achieves 12.0\% ASR, a 66\% reduction from the 35.6\% baseline. The three stronger models also reduce ASR but by much less. Claude 4.5 Sonnet and GPT-5 reach about 22\%, and Gemini-3-pro only drops to 28.8\%. Answer length does not explain this gap. GPT-5 and GPT-4o-mini produce answers of similar length, but GPT-5's ASR is nearly twice as high. We looked at the image prompts from each model and found the real difference there. GPT-4o-mini writes short, concrete prompts. They focus on specific visual elements like dark lighting, red warning signs, and surveillance cameras. Doubao receives clear instructions from these prompts and produces images with strong, consistent threat atmosphere. The concise answers from GPT-4o-mini also train the student model to respond briefly. This helps the model produce short refusals at test time. The stronger models produce noisier prompts. GPT-5 tends toward abstract, high-level descriptions. Claude 4.5 Sonnet tries to balance threatening and reassuring elements in the same prompt. Gemini-3-pro packs too many visual details into each prompt. In all three cases, Doubao receives unfocused instructions and produces images with weaker threat signals. This pattern is consistent with observations in knowledge distillation research. Gu et al. \cite{gu2024minillm} showed that the largest model is not always the best teacher, and smaller models can sometimes produce more focused training signals. Based on this comparison, we selected GPT-4o-mini as the teacher model for all experiments in the main paper. Among the four models we tested, it produces the clearest threat imagery and the most effective training data.

\section{Visual Style Ablation}
\label{app:style}

This appendix examines whether the safety effect of VSFA depends on any particular visual style. Our main experiments use a mix of 12 visual styles when generating threat-related images. To isolate the contribution of each style, we trained Qwen2.5-VL-7B separately with 50 images of each style. We generated the corresponding VQA pairs through the same pipeline described in Section 3. We used LoRA rank 128 and only fine-tuned the language model. The visual encoder stayed frozen. We evaluated each model on FigStep with GPT-4o as the judge. Table \ref{tab:style_ablation} shows the results.

\begin{table}[ht]
\centering
\caption{Visual style ablation on Qwen2.5-VL-7B. Each row is a separate training run using 50 images of one style. Lower ASR means safer.}
\label{tab:style_ablation}
\small
\begin{tabular}{lcc}
\toprule
\textbf{Visual Style} & \textbf{Images} & \textbf{FigStep ASR (\%) $\downarrow$} \\
\midrule
No Defense (baseline) & 0 & 35.6 \\
\midrule
Photorealistic & 50 & 12.8 \\
Digital art & 50 & 13.1 \\
Concept art & 50 & 12.4 \\
Technical illustration & 50 & 13.7 \\
Documentary style & 50 & 11.9 \\
Abstract representation & 50 & 12.6 \\
Cinematic & 50 & 11.4 \\
Artistic & 50 & 11.2 \\
Professional & 50 & 12.1 \\
Casual & 50 & 13.1 \\
Futuristic & 50 & 12.8 \\
Vintage & 50 & 13.3 \\
\bottomrule
\end{tabular}
\end{table}

All 12 styles achieve ASR between 11.2\% and 13.7\%. The mean is 12.5\% and the standard deviation is only 0.74\%. Even the highest and lowest ASR only differ by 2.5 percentage points. Photorealistic and Abstract representation are the two most visually different styles in our set. One looks like a photograph. The other is stylized and non-literal. Yet their ASR results are almost the same , a gap of just 0.2 percentage points. This means the safety effect does not depend on visual style. The alignment signal comes from what the image depicts, not from how it looks. Our SAE analysis in Appendix~C supports this. The most activated latent encodes abstract safety concepts rather than visual style features. Our finding also has a practical benefit. Practitioners do not need to optimize for any particular visual style when building VSFA training sets. It also explains why VSFA generalizes to attack categories not seen during training. The safety persona is activated by the semantic content of threat-related scenes, regardless of how they are rendered.

\section{General Capability Evaluation (MMLU and MMMU)}
\label{app:mmlu}

This appendix reports VSFA's performance on two general reasoning benchmarks: MMLU \cite{hendrycks2021mmlu} and MMMU \cite{yue2024mmmu}. MMLU covers 57 subjects across STEM, humanities, social sciences, and other domains. MMMU covers 30 subjects with college-level questions that require both visual and textual reasoning. We evaluated Qwen2.5-VL-7B before and after VSFA training using the standard protocol for both benchmarks. Table \ref{tab:mmlu_mmmu} shows the results.

\begin{table}[ht]
\centering
\caption{General capability evaluation on Qwen2.5-VL-7B before and after VSFA training.}
\label{tab:mmlu_mmmu}
\small
\begin{tabular}{lccc}
\toprule
\textbf{Benchmark} & \textbf{Baseline} & \textbf{VSFA} & \textbf{Diff} \\
\midrule
MMLU & 68.52\% & 68.13\% & $-$0.39\% \\
MMMU & 50.56\% & 50.11\% & $-$0.45\% \\
\bottomrule
\end{tabular}
\end{table}

MMLU drops by 0.39\% and MMMU drops by 0.45\%. Both drops are well under 0.5\%, which is within the normal variance of safety fine-tuning methods. Zong et al. \cite{zong2024vlguard} reported MMLU changes between $-$0.08\% and +1.79\% across different VLGuard configurations. Our drops are at the lower end of this range. Combined with MM-Vet results in Table 2, VSFA preserves general capabilities across MMLU, MMMU, and MM-Vet.

\section{Cross-Family Generalization}
\label{app:cross}

This appendix tests whether VSFA generalizes beyond the 7B to 8B models used in the main paper. Our main experiments cover four models from two families: Qwen2.5-VL-7B, Qwen3-VL-8B, LLaVA-1.5-7B, and LLaVA-v1.6-Mistral-7B. To test across different scales and architectures, we applied VSFA to two additional models: Gemma 3 IT (4B) \cite{team2025gemma} and Llama 3.2 Vision (11B) \cite{grattafiori2024llama3}. Gemma 3 IT uses a 4B language backbone with a SigLIP-based vision encoder. Llama 3.2 Vision uses an 11B backbone with a different vision-language integration architecture. Neither shares a vision encoder, projection layer, or LLM backbone with the Qwen or LLaVA families. We ran the standard VSFA pipeline on each model with only 60 training images, fine-tuned with LoRA under identical settings, and evaluated on FigStep with GPT-4o as the judge. Table \ref{tab:cross_family} shows the results.

\begin{table}[ht]
\centering
\caption{VSFA on additional model families. Each model is trained with only 60 images. Lower ASR means safer.}
\label{tab:cross_family}
\small
\resizebox{\columnwidth}{!}{%
\begin{tabular}{lcccc}
\toprule
\textbf{Model} & \textbf{Size} & \textbf{Baseline ASR} & \textbf{VSFA ASR} & \textbf{Diff} \\
\midrule
Gemma 3 IT & 4B & 37.0\% & 7.2\% & $-$29.8\% \\
Llama 3.2 Vision & 11B & 49.8\% & 17.6\% & $-$32.2\% \\
\bottomrule
\end{tabular}
}
\end{table}

Gemma 3 IT drops from 37.0\% to 7.2\% ASR, and Llama 3.2 Vision drops from 49.8\% to 17.6\%. Both are large reductions with only 60 training images. All six models now tested use different vision encoders and LLM backbones, covering sizes from 4B to 11B. VSFA produces consistent ASR reductions across all of them. This suggests that the mechanism is not tied to any particular architecture or model size. It is consistent with our SAE finding (Appendix~C) that VSFA activates general safety-oriented persona features that appear to exist across different model families. The data efficiency is also notable. Our main experiments use 700 images, but here we used only 60 and still achieved large ASR reductions. This confirms that VSFA does not require large-scale training data to be effective.

\end{document}